\documentclass[10pt,conference]{IEEEtran}
\IEEEoverridecommandlockouts
\usepackage{cite}
\usepackage{amsmath,amssymb,amsfonts}
\usepackage{algorithmic}
\usepackage{graphicx}
\usepackage{textcomp}
\usepackage{xcolor}
\usepackage{xspace}
\usepackage[export]{adjustbox}
\usepackage{multirow}

\makeatletter
\DeclareRobustCommand\onedot{\futurelet\@let@token\@onedot}
\def\@onedot{\ifx\@let@token.\else.\null\fi\xspace}
 
\def\ie{\emph{i.e}\onedot}

\def\etal{\emph{et al}\onedot}
\makeatother

\usepackage{rotating} 

\newcommand{\bI}{\mathbf{I}}

\begin{document}

\title{Diffusion-based image inpainting with internal learning}

\author{\IEEEauthorblockN{Nicolas Cherel}
\IEEEauthorblockA{\textit{LTCI, Télécom Paris} \\
\textit{IP Paris}}
\and
\IEEEauthorblockN{Andrés Almansa}
\IEEEauthorblockA{\textit{MAP5, CNRS \& Université Paris Cité}}
\and
\IEEEauthorblockN{Yann Gousseau}
\IEEEauthorblockA{\textit{LTCI, Télécom Paris} \\
\textit{IP Paris}}
\and
\IEEEauthorblockN{Alasdair Newson}
\IEEEauthorblockA{\textit{ISIR, Sorbonne Université} }
}

\maketitle

\begin{abstract}
Diffusion models are now the undisputed state-of-the-art for image generation and image restoration. However, they require large amounts of computational power for training and inference.
In this paper, we propose lightweight diffusion models for image inpainting that can be trained on a single image, or a few images. We show that our approach competes with large state-of-the-art models in specific cases.
We also show that training a model on a single image is particularly relevant for image acquisition modality that differ from the RGB images of standard learning databases.
We show results in three different contexts: texture images, line drawing images, and materials BRDF, for which we achieve state-of-the-art results in terms of realism, with a computational load that is greatly reduced compared to concurrent methods.
\end{abstract}


\section{Introduction}

Inpainting, or image completion, is the task of filling in an unknown zone in an image in a manner which is coherent with the surrounding context of this image. Some of the first methods for this task include exemplar-based methods~\cite{criminisiRegionFillingObject2004,wexlerSpaceTimeCompletionVideo2007,newsonNonLocalPatchBasedImage2017}.
These techniques rely on the powerful self-similarity property of natural images to fill in the occlusion.
They are particularly good at synthesizing convincing textures that fit in the image. 

Deep learning, on the other hand, takes the approach of training a model on a large external dataset, leading to impressive results. Assuming similarity between the training data and testing data, the network can thus efficiently inpaint missing content~\cite{pathak_context_2016}.
The advantage of this approach, compared to patch-based methods, is that it can reconstruct semantic content that is not found in the test data. This is particularly useful, for example, in the case of facial images where the relevant content (eyes, mouth) is hidden by the occlusion.

Even more recently, new incredible results have been produced by diffusion models~\cite{hoDenoisingDiffusionProbabilistic2020b}, surpassing previous generative approaches like Generative Adversarial Networks (GANs).

However, these require extremely large models, making the approach computationally expensive. The central reason why these models are so large is that many different sub-tasks must be carried out during diffusion-based inpainting. At the beginning of the diffusion, the global geometry of the image must be reconstructed, followed by finer details at the end of the diffusion.

Training large models on large datasets incontestably yields great results, however, in this work we show that these ``external'' methods, trained on external data, can be matched by internal methods, which use only the test data or very limited training data.
In this context, we propose to use a lightweight neural network based on the diffusion framework for the diverse inpainting of images. Our neural network is trained only on the test image, using the available visible data.
We thus overfit on this single example to have a specialized model. 
We also show that in specific contexts such as material inpainting, there is no existing solution available off-the-shelf, and an internal method is well-suited given the specifics of such modalities, and the limited availability of large datasets.
On material BRDF datasets, our model produces very good results and is perfectly adapted to the specificities of these images, maintaining the correlation between the different maps.

\section{Related works}

Classical methods for image inpainting have long relied on the self-similarity of natural images, building solutions on patches from the inpainted image. Two classical approaches include the greedy exemplar-based method of Criminisi~\etal~\cite{criminisiRegionFillingObject2004} and the global optimization method by Wexler~\etal~\cite{wexlerSpaceTimeCompletionVideo2007}. Several improvements have been proposed to make this last approach faster (Barnes~\etal~\cite{barnesPatchMatchRandomizedCorrespondence2009b}) or to include texture features (Newson~\etal~\cite{newsonNonLocalPatchBasedImage2017}).

With the development of deep learning, it has become common to train a neural network on a large database of images and to perform the inpainting with a simple forward pass.
The seminal work of Pathak~\etal~\cite{pathak_context_2016} has been refined with global and local discriminators by Iizuka~\etal~\cite{iizuka_globally_2017}. Yu~\etal~\cite{yuGenerativeImageInpainting2018} adds an attention layer which mimicks the classical patch-based approaches.
The method LAMA~\cite{suvorovResolutionrobustLargeMask2022} uses Fourier convolutions to avoid expensive attention layers.

Recently, diffusion models \cite{hoDenoisingDiffusionProbabilistic2020b}, acclaimed for their generative performances, have been used for image inpainting. Lugmayr~\etal~\cite{lugmayrRePaintInpaintingUsing2022} derive an inpainting algorithm from a purely unconditional generative model. Saharia~\etal~\cite{sahariaPaletteImagetoImageDiffusion2022b} train a diffusion model specifically for image inpainting. These approaches enable state-of-the-art inpainting results but require massive training procedures. 

A series of work have explored the training of deep networks on very small datasets and even single images, we refer to these methods as internal learning. Deep Image Prior~\cite{ulyanovDeepImagePrior2018a} has been proposed to use a neural network as the regularization component for various image restoration methods.
Following SinGAN~\cite{shaham2019singan} for single image generation, Alkobi~\etal~\cite{alkobiInternalDiverseImage2023} proposed a network trained on a single image for image inpainting. It can generate diverse image completions.
A preliminary version of our work was presented in \cite{cherelModeleDiffusionFrugal2023}.

\section{Method}

\subsection{Framework}
We adapt the general framework of diffusion models~\cite{hoDenoisingDiffusionProbabilistic2020b} to image inpainting.
Starting from the unknown distribution of natural images $q(x_0 | y)$ conditioned on the observations $y$, we define a Markov chain $q(x_0 | y), ..., q(x_T | y)$ with the forward transition kernel for $t \in [1,T]$:

\begin{equation}
    q(x_{t} | x_{t-1}, y) = \mathcal{N}\left( x_t; \sqrt{1 - \beta_t} x_{t-1},  \beta_t \bI \right),
\end{equation}
with the variance schedule linearly increasing from $\beta_1=1\mathrm{e}{-4}$ to $\beta_T=0.02$.
We introduce the variables $\alpha_t = 1 - \beta_t$ and $\bar{\alpha}_t = \prod_{s=1}^t \alpha_s$ which will come in handy because $q(x_{t} | x_{0}) = \mathcal{N}\left( x_t; \sqrt{\bar\alpha_t} x_{0},  (1-\bar\alpha_t) \bI \right)$.
In this case, the reverse process defines a conditional Gaussian distribution $p_\theta(x_t | x_{t+1}, y)$ of learnable mean and variance:

$$
p_\theta \left( x_t | x_{t+1}, y \right) = \mathcal{N}\left( x_t; \mu_\theta(x_{t+1},y,t), \sigma^2_t\bI \right)
$$

As is the standard approach, we use a neural network (detailed below) to predict the conditional mean $\mu_\theta(x_{t+1}, y, t)$. We use a fixed variance schedule for the reverse process: $\sigma^2_t =  \frac{1 - \bar{\alpha}_{t-1}}{1 - \bar{\alpha}_t} \beta_t$.
The parameters $\theta$ are optimized to maximize a lower bound of the log-likelihood of the data in \cite{hoDenoisingDiffusionProbabilistic2020b}.

In the case of inpainting, as mentioned above, we are interested in the data distribution conditioned on our observations \ie the known region of the image.
The observations are introduced in the network as incomplete images via concatenation in the input layer (similarly to \cite{sahariaPaletteImagetoImageDiffusion2022b}).
We also add the mask information as input.
We train our network by minimizing the reweighted $\ell_2$ loss of the $x_0$-parametrization:

\begin{equation}
  \label{eq:denoising}
    \mathbb{E}_{x_0, \epsilon, t, M} \lVert M \odot \left( x_0 - f_\theta(x_t, y, t, M) \right) \rVert_2^2,
\end{equation}
where $M \sim \mathcal{P}_\text{mask}$ is a binary mask with $0$ for the known region and $1$ elsewhere, $y$ is the masked image \ie $y = x_0 \odot (1 - M)$, and $x_t$ is obtained by the forward diffusion of $x_0$ by the following:
$$ x_t = \sqrt{\bar\alpha_t} x_0 + \sqrt{1 - \bar\alpha_t} \epsilon $$
where $\epsilon \sim \mathcal{N}(0, \bI)$.

The parameter $t$ controls the noise level. Our network $f_\theta$ takes as input the noisy image $x_t$, the observation $y$, the mask $M$ and the noise level at step $t$.

\subsection{Single image inpainting}
In the case of single image inpainting, we are given a large image $x_\text{test}$ and its corresponding mask $M_\text{test}$. When possible, we take sub-regions of the large image that do not intersect with the testing mask, and train on these smaller images: we generate synthetic masks, pick a random timestep, and sample noise to be added.
After training, the model is then applied to the test data.

When the image is too small to take sub-regions, we still use the same training tactic but mask the input data with both the training mask and the testing mask. The loss for the pixels in the testing mask is zero.

\subsection{Architecture}
\begin{figure}
    \centering
    \includegraphics[width=0.45\textwidth]{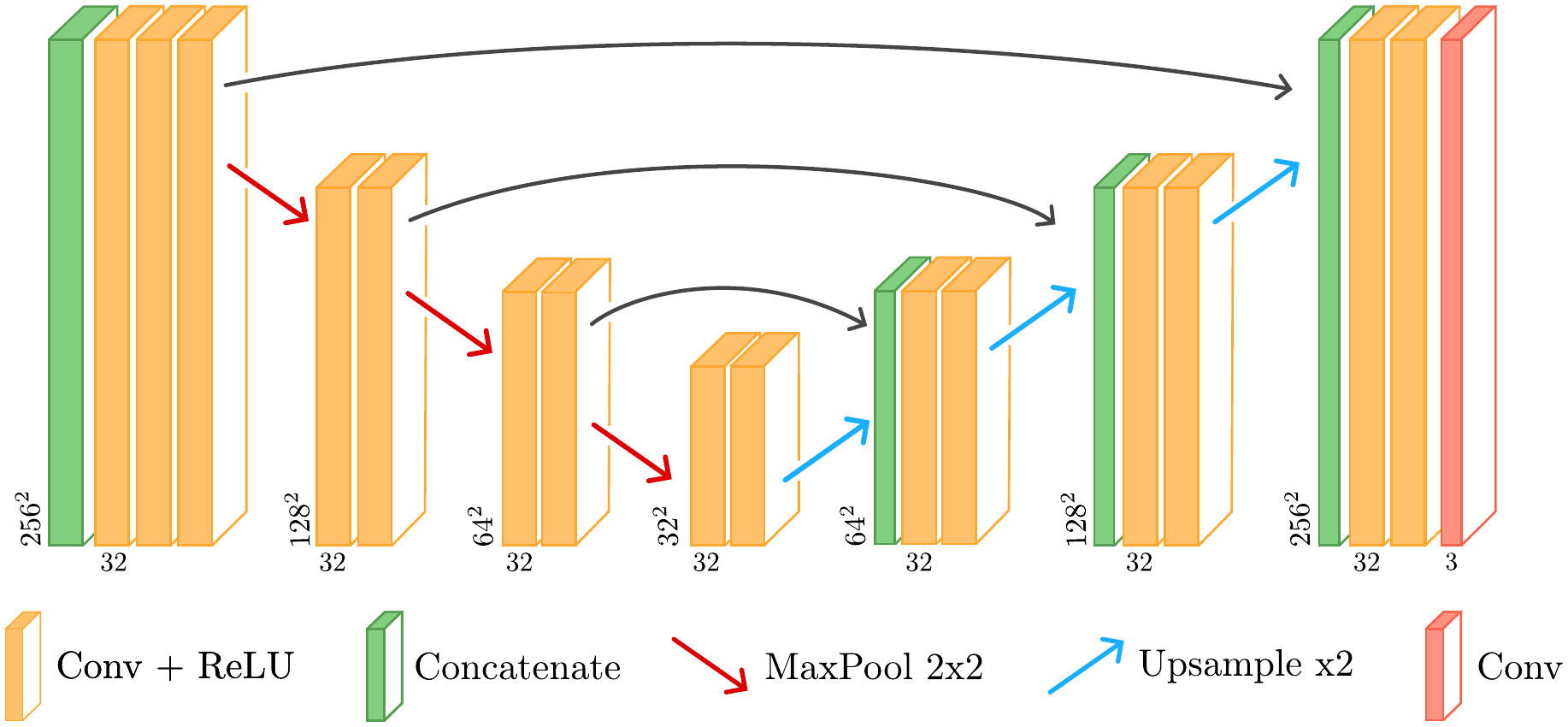}
    \caption{UNet architecture used for our experiments. The different inputs are first concatenated before being processed by fully convolutional layers.}
    \label{fig:unet}.
\end{figure}

In practice, our conditional network takes as input the masked image $y$, the noisy image $x_t$, the inpainting binary mask $M$, and the temporal information $t$.
We use a much smaller network architecture than in the classical approach \cite{hoDenoisingDiffusionProbabilistic2020b}.
We also use a UNet-type network, but without attention layers. The network entries are concatenated before the first convolution and the number of channels is limited to 32.
As a result, our network has 160k parameters, compared to 450M for the state-of-the-art RePaint method~\cite{lugmayrRePaintInpaintingUsing2022}.
The architecture is represented in Figure~\ref{fig:unet}.

Contrary to \cite{hoDenoisingDiffusionProbabilistic2020b}, we get better results by directly predicting the initial image $x_0$ rather than the noise $\epsilon$.
We find that our simple formulation, similar to that used in some denoising networks, gives us equivalent results and simplifies the network.
Training is performed by randomly drawing regions of size 256x256 from our reference image, which are then noised and masked.
The $M$ mask is generated synthetically as in \cite{yuGenerativeImageInpainting2018}. The algorithm is designed to mimick an eraser used for removing a stain with back and forth motions in a local region. Starting from a random starting point, a sequence of lengths and angles are drawn randomly and defines the brushstrokes. This is repeated multiple times.

It is necessary to train the network to perform denoising for all time steps, so they are drawn uniformly in $[1,T]$.

The UNet architecture allows for an adaptive receptive field: depending on the task, the multi-scale nature of the network can be exploited or not. This is a clear advantage over patch-based methods, for which the patch size and the number of scales in the Gaussian pyramid are two very important parameters that can be difficult to set.

\section{Experiments}

\subsection{Methods}
For all experiments, we train our diffusion model for 15k iterations for each image. The initial learning rate is 1e-4 and is lowered to 1e-5 after 10k iterations. Training time is 15 minutes on a NVidia V100 GPU.
We compare our method to the internal patch-based method of Newson~\etal~\cite{newsonNonLocalPatchBasedImage2017}.
We also include comparisons with a very large, state-of-the-art external diffusion model trained on natural images for image inpainting: RePaint~\cite{lugmayrRePaintInpaintingUsing2022}.
Our method can also produce diverse solutions but we do not evaluate this property here. 

\subsection{Textures}

In order to show the capacity of the method to leverage internal regularity, we test it on 100 images from the texture dataset of \cite{linUniversalTextureSynthesis2023}. Many of these images have self-similar structures and are well suited for internal learning and patch-based methods.

\begin{figure*}
  \centering
  \setlength{\tabcolsep}{0.9pt}
  \begin{tabular}{cccccc}
    \raisebox{0.09\textwidth}{\rotatebox[origin=c]{90}{Patch}}
    & \includegraphics[width=0.19\textwidth]{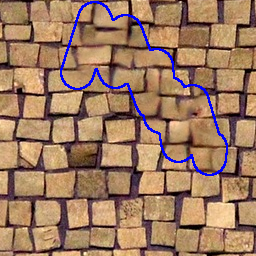}
    & \includegraphics[width=0.19\textwidth]{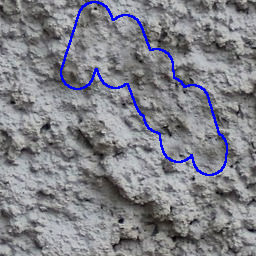}
    & \includegraphics[width=0.19\textwidth]{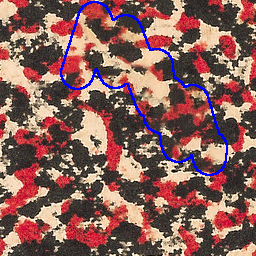}
    & \includegraphics[width=0.19\textwidth]{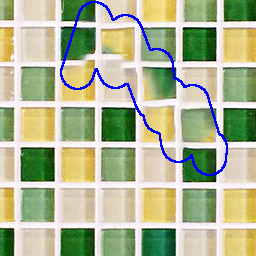}
    & \includegraphics[width=0.19\textwidth]{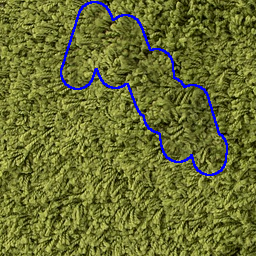}
    \\
    \raisebox{0.09\textwidth}{\rotatebox[origin=c]{90}{DeepFill}}
    & \includegraphics[width=0.19\textwidth]{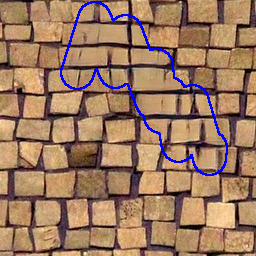}
    & \includegraphics[width=0.19\textwidth]{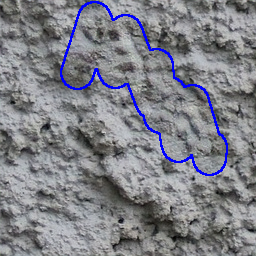}
    & \includegraphics[width=0.19\textwidth]{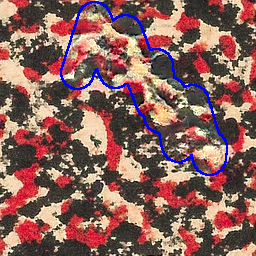}
    & \includegraphics[width=0.19\textwidth]{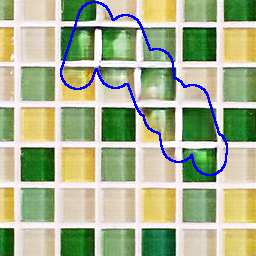}
    & \includegraphics[width=0.19\textwidth]{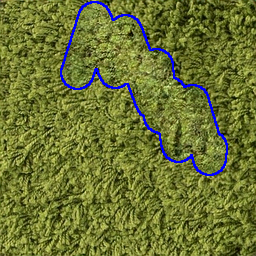}
    \\
    \raisebox{0.09\textwidth}{\rotatebox[origin=c]{90}{RePaint}}
    & \includegraphics[width=0.19\textwidth]{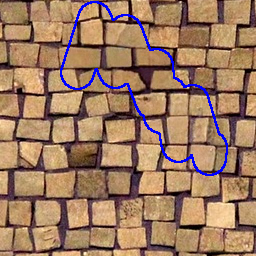}
    & \includegraphics[width=0.19\textwidth]{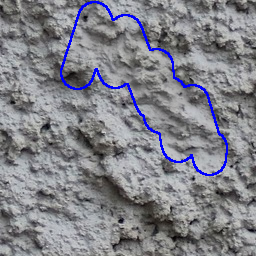}
    & \includegraphics[width=0.19\textwidth]{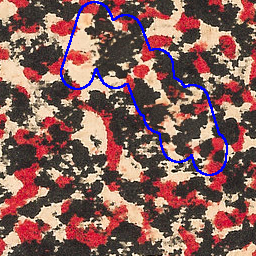}
    & \includegraphics[width=0.19\textwidth]{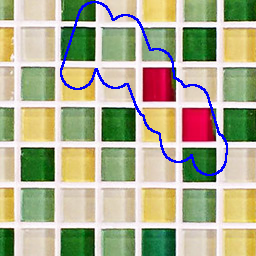}
    & \includegraphics[width=0.19\textwidth]{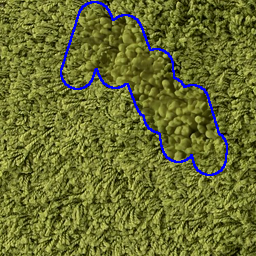}
    \\
    \raisebox{0.09\textwidth}{\rotatebox[origin=c]{90}{Ours}}
    & \includegraphics[width=0.19\textwidth]{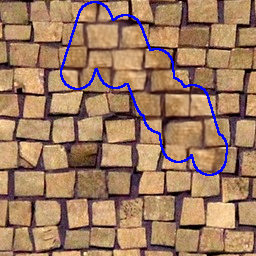}
    & \includegraphics[width=0.19\textwidth]{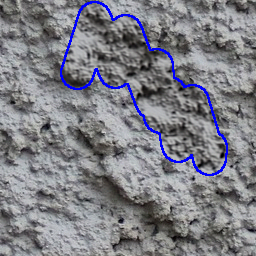}
    & \includegraphics[width=0.19\textwidth]{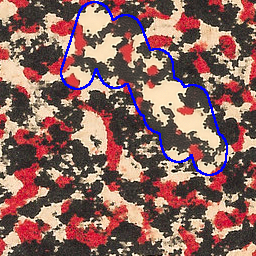}
    & \includegraphics[width=0.19\textwidth]{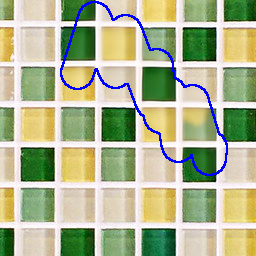}
    & \includegraphics[width=0.19\textwidth]{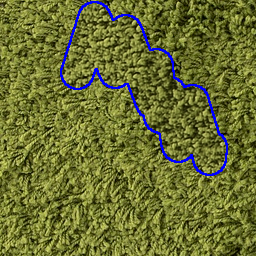}
    \end{tabular}
    \caption{Inpainting results for 4 methods (see text). Results from DeepFill are the least satisfying. Results from RePaint are good but sometime lacks sharpness, produce results with wrong scales or hallucinate content (4th column). Patch and our method yield the best results in these cases.}
    \label{fig:textures}
  \end{figure*}

We show  some visual results in Figure~\ref{fig:textures}. We see that the diffusion model produces very crisp results, on par with the results from RePaint. 
RePaint may suffer from hallucination and bias from the external datasets despite performing very well on many challenging textures. This can be seen with the appearance of red squares in one of the examples. 
We measure the inpainting error with PSNR, and LPIPS which are not optimal given the ill-defined nature of inpainting but reasonable for textures (a lot less for non-stationary images with many degrees of freedom) in Table~\ref{tab:textures}.
We also include a comparison with a lighter external method, DeepFill~\cite{yuGenerativeImageInpainting2018}. This approach  produces relatively poor results in comparisons to the others. 
RePaint produces very good results, at the cost of a huge number of parameters (3 orders of magnitude more than our approach).
The internal methods, patch-based and ours, yield the best performances on this database.

\begin{table}
  \centering
  \caption{Reconstruction error using different metrics on our texture inpainting dataset.}
  \begin{tabular}{|c|c|c|c|c|c|}
    \hline
    & Method & PSNR (dB) ↑ & SSIM ↑ & LPIPS ↓  & \#Param.\\
    \hline
    \multirow{2}{*}{External} & DeepFill &  21.55 & 0.870 & 0.105 & 3M\\
    & RePaint &  26.77 & 0.899 & 0.053 &  450M\\
    \hline
    \multirow{2}{*}{Internal} & Patch &  \textbf{28.12} & 0.902 & \textbf{0.046} &  -\\
    & Ours &  28.02 & \textbf{0.908} & 0.057 &  160k\\
    \hline
    \end{tabular}
    \label{tab:textures}
\end{table}

\subsection{Line drawing}

\begin{figure}
  \centering
  \setlength{\tabcolsep}{0.9pt}
  \begin{tabular}{ccc}
    \includegraphics[width=0.16\textwidth,cfbox=black 1pt -1pt]{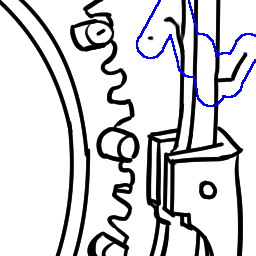}
    & \includegraphics[width=0.16\textwidth,cfbox=black 1pt -1pt]{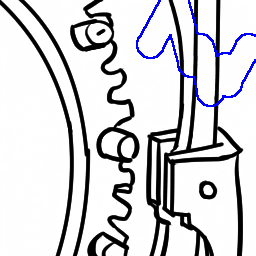}
    & \includegraphics[width=0.16\textwidth,cfbox=black 1pt -1pt]{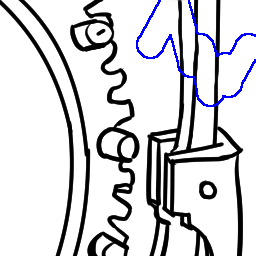}
    \\
     \includegraphics[width=0.16\textwidth,cfbox=black 1pt -1pt]{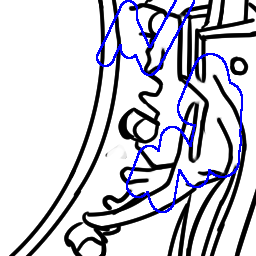}
    & \includegraphics[width=0.16\textwidth,cfbox=black 1pt -1pt]{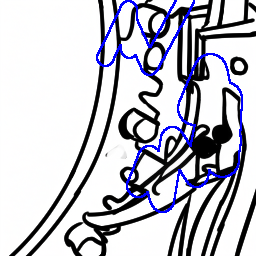}
    & \includegraphics[width=0.16\textwidth,cfbox=black 1pt -1pt]{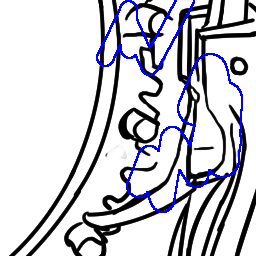}
    \\
    Patch & RePaint & Ours
    \end{tabular}
    \caption{For simple completions (top), our method performs as well as RePaint and better than the method of Newson~\etal. For complex completions with no obvious ground-truth, all results are good.}
    \label{fig:line_drawing}
  \end{figure}

We also include an evaluation on line drawing inpainting, using the dataset  from Sasaki~\etal~\cite{sasakiLearningRestoreDeteriorated2018}.
Our method is trained on the corresponding training set. We generate a testing set using subregions of the test images (100 of them).
For the patch-based method of Newson~\etal, we only use the reference image as there is no easy way to use the rest of the training set.
We show some results in Figure~\ref{fig:line_drawing}.
The results are visually very good for all three methods, with sharp lines and no blurring.
We measure the results quantitatively using binary cross-entropy which is well suited to binary modalities and the LPIPS metric.
For this task, we show that our method is competitive with RePaint (Table~\ref{tab:line_drawing}), but with only a fraction of the inference time (3 seconds for our method compared to 10 minutes for RePaint). The diversity, speed, and quality of results make our method interesting for interactive uses.

\begin{table}
  \centering
  \caption{Reconstruction error using different metrics on line drawing}
  \begin{tabular}{|c|c|c|c|c|}
    \hline
    Method & Cross-entropy ↓ & LPIPS ↓  & Training time & Inference time\\
    \hline
    Patch & 0.269 & 0.110  &  - & 2 min\\
    RePaint & 0.113 & \textbf{0.042} &  24 days & 10 min\\
    Ours & \textbf{0.080} & 0.047 &  15 min & 3 sec \\
    \hline
    \end{tabular}
    \label{tab:line_drawing}
\end{table}

\subsection{SVBRDF}

Spatially Varying Bidirectional Reflectance Distribution Functions (SVBRDF) are a common representation of materials used to render textures on 3D objects. These spatial functions encode the appearance and physical properties of each material to model their interaction with light for rendering purposes.
These maps are highly correlated, and their inpainting must be done jointly.
We use 50 examples from the Deschaintre~\etal~\cite{deschaintreSingleImage} dataset, which includes the diffuse color map, the normal map, the roughness map and the specular map. Other types of maps (metallic, displacement, etc.) may be included depending on the material, which complicates the development of a single inpainting model. 
SVBRDFs are used to decorate objects and structures in 3D worlds, and have similar properties to image textures. An internal method thus appears well-suited to inpaint SVBRDFs.

We adapt the patch-based method of Newson~\etal to use more than 3 input channels and to reconstruct all channels together, simply by modifying the distance used to compare patches.
For our method, we simply add more input and output channels. We do not reweight the channels.
Finally, we compare with an external method (RePaint), which is used to inpaint each map independently. In details, we perform 4 separate inpaintings for the diffuse color, normal map, roughness map, and specular map. Observe that the retraining of the corresponding diffusion model would imply massive training times and very large databases. 

\begin{figure*}
    \centering
    \setlength{\tabcolsep}{1.5pt}
    \small
    \begin{tabular}{cccccc}
        \raisebox{0.09\textwidth}{\rotatebox[origin=c]{90}{Patch}} &\includegraphics[width=0.18\textwidth]{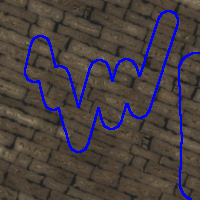}
        & \includegraphics[width=0.18\textwidth]{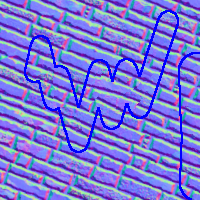}
        & \includegraphics[width=0.18\textwidth]{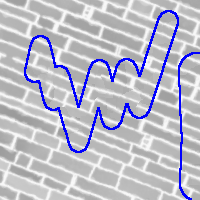}
        & \includegraphics[height=0.18\textwidth]{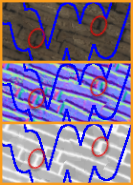}
        & \includegraphics[height=0.18\textwidth]{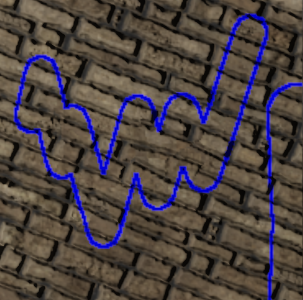} \\
        \raisebox{0.09\textwidth}{\rotatebox[origin=c]{90}{RePaint}} & \includegraphics[width=0.18\textwidth]{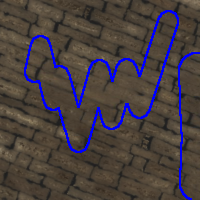}
        & \includegraphics[width=0.18\textwidth]{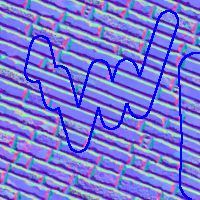}
        & \includegraphics[width=0.18\textwidth]{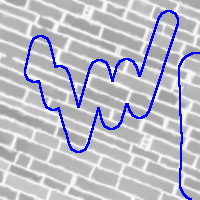}
        & \includegraphics[height=0.18\textwidth]{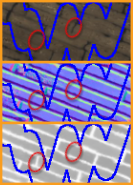} 
        & \includegraphics[height=0.18\textwidth]{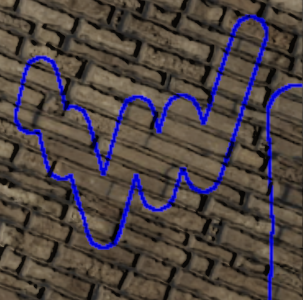} \\
        \raisebox{0.09\textwidth}{\rotatebox[origin=c]{90}{Ours}} &\includegraphics[width=0.18\textwidth]{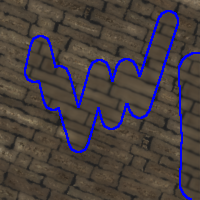}
        & \includegraphics[width=0.18\textwidth]{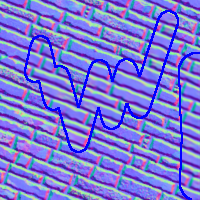}
        & \includegraphics[width=0.18\textwidth]{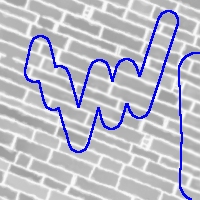}
        & \includegraphics[height=0.18\textwidth]{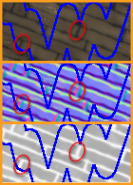} 
        & \includegraphics[height=0.18\textwidth]{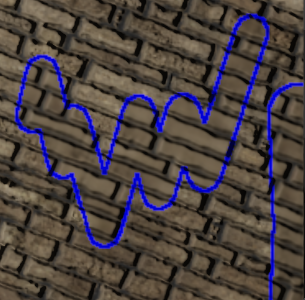} \\
        & Diffuse & Normal & Roughness & Zoom-In & Render
    \end{tabular}
    \caption{Inpainting results for SVBRDFs. The properties of the materials are correlated across maps, so inpainting should maintain this correlation. Our method preserves this property while non-specific methods like RePaint does not (see zoom-in with borders circled in red, where the short lines are not in coherent positions). RePaint's render looks flat. The specular map is omitted as this material has a uniform specularity which is correctly inpainted by all approaches.}
    \label{fig:materials}
\end{figure*}

The results are shown in Figure~\ref{fig:materials}.  It can be seen that RePaint produces incoherent completions despite good individual inpainting. On the other hand, the patch-based method and our internal method preserve much better the correlations between maps. 
This is important for rendering the materials as seen in the last column. The lack of correlation produces results which lacks simulated depth, which is the original goal of SVBRDF.
Quantitatively, we simply measure the mean square error between the ground truth and the reconstruction. A better evaluation would be to evaluate the renderings in different lighting conditions, but that is beyond the scope of this work.
We see in Table~\ref{tab:materials} that the MSE is lower for our method.

\begin{table}
    \centering
    \caption{Quantitative evaluation of material inpainting for all the maps using mean-squared error.}
    \begin{tabular}{|c|c|c|c|c|}
    \hline
    Method & Diffuse ↓ & Normals ↓ & Roughness ↓ & Specular ↓ \\
    \hline
    RePaint & 0.0024 & 0.0058 & 0.0069 & 0.00021 \\
    Patch & 0.0031 & 0.0071 & 0.0078 & 0.00020 \\
    Ours & \textbf{0.0017} & \textbf{0.0050} & \textbf{0.0045} & \textbf{0.00012} \\
    \hline
    \end{tabular}
    \label{tab:materials}
\end{table}

\section{Conclusion}

In this paper, we have shown that a lightweight diffusion model trained on a single image or a few images is very efficient for inpainting self-similar images. In such cases, results are on par or better than results produced by massive diffusion models implying training or inference times that are several order of magnitude larger. As a result, our method can be used in a variety of contexts to achieve high-quality and diverse inpainting results.
this would be particularly interesting for interactive use.
We have shown on material data that image-like data can be handled much better by specific diffusion models than by large standard networks. This is of great interest in cases where data availability is low and the training of large networks is infeasible.

Our approach could benefit from the most common techniques for speeding up diffusion models, as well as any quantization or pruning technique, to make it lighter and faster.

\section*{Acknowledgements}

The authors acknowledge support from the French Research
Agency through the PostProdLEAP project (ANR-19-CE23-0027-01) and Mistic project (ANR-19-CE40-005).

\bibliographystyle{IEEEtran}
\bibliography{IEEEabrv,references}

\end{document}